\documentclass[]{spie}  

\usepackage{amsmath,amsfonts,amssymb}
\usepackage{graphicx}
\usepackage[colorlinks=true, allcolors=blue]{hyperref}

\title{Multi-scale Sparse Representation-Based Shadow Inpainting for Retinal OCT Images}

\author[a]{Yaoqi Tang}
\author[a]{Yufan Li}
\author[b]{Hongshan Liu}
\author[a]{Jiaxuan Li}
\author[c]{Peiyao Jin}
\author[b]{Yu Gan}
\author[a,*]{Yuye Ling}
\author[d]{Yikai Su}
\affil[a]{\small{John Hopcroft Center for Computer Science, Shanghai Jiao Tong University, Shanghai 200240, China}}
\affil[b]{Department of Electrical and Computer Engineering, The University of Alabama, AL 35487, USA}
\affil[c]{Department of Ophthalmology, Shanghai General Hospital, Shanghai Jiao Tong University, Shanghai 200080, China}
\affil[d]{State Key Lab of Advanced Optical Communication Systems and Networks, Department of Electronic
Engineering, Shanghai Jiao Tong University, Shanghai 200240, China}

\authorinfo{$^*$Send correspondence to: yuye.ling@sjtu.edu.cn}

\begin{document} 
\maketitle

\begin{abstract}
Inpainting shadowed regions cast by superficial blood vessels in retinal optical coherence tomography (OCT) images is critical for accurate and robust machine analysis and clinical diagnosis. Traditional sequence-based approaches such as propagating neighboring information to gradually fill in the missing regions are cost-effective. But they generate less satisfactory outcomes when dealing with larger missing regions and texture-rich structures. Emerging deep learning-based methods such as encoder-decoder networks have shown promising results in \emph{natural} image inpainting tasks. However, they typically need a long computational time for network training in addition to the high demand on the size of datasets, which makes it difficult to be applied on often small \emph{medical} datasets. To address these challenges, we propose a novel multi-scale shadow inpainting framework for OCT images by synergically applying sparse representation and deep learning: sparse representation is used to extract features from a small amount of training images for further inpainting and to regularize the image after the multi-scale image fusion, while convolutional neural network (CNN) is employed to enhance the image quality. During the image inpainting, we divide preprocessed input images into different branches based on the shadow width to harvest complementary information from different scales. Finally, a sparse representation-based regularizing module is designed to refine the generated contents after multi-scale feature aggregation. Experiments are conducted to compare our proposal versus both traditional and deep learning-based techniques on synthetic and real-world shadows. Results demonstrate that our proposed method achieves
favorable image inpainting in terms of visual quality and quantitative metrics, especially when wide shadows are presented. 
\end{abstract}


\section{INTRODUCTION}
\label{sec:intro}  

Retinal optical coherence tomography (OCT) images are widely used in clinical settings for diagnosing ophthalmic diseases \cite{2017Retinal}. Despite of the $\mu$m-level resolution and the high sensitivity manifested in OCT images, artifacts are often spotted including the shadows that are caused by scattering from the superficial retinal vessels. These artifacts would decrease the accuracy of automated segmentation algorithms and cause errors in clinical diagnosis \cite{Li:21}. Therefore, an accurate shadow inpainting algorithm that is applicable for suppressing such artifacts in retinal OCT images is of great interest to the community.

Inpainting shadow regions while simultaneously maintaining overall consistency for retinal OCT images could be categorized as an image restoration problem, which attributes to many practical applications and has been extensively advanced in the field of computer vision \cite{2019Image}. Traditional sequence-based methods \cite{2010Image} \cite{2010Wavelet} \cite{2011Shadow}for this task that gradually fill in the missing area by diffusing neighboring information or searching for most similar patches from known regions are widely utilized. But they cannot handle complicated structures and fail to recover realistic results due to the lack of global information. Sparse representation-based methods \cite{article2019} can extract features of images by decomposing data into a low-dimensional space and has shown great effectiveness in various image processing tasks \cite{7102696}. Besides, recent research efforts turn to deep learning-based techniques \cite{2019Deep}. For example, Li \emph{et al.} proposed a CNN-based method using Recurrent Feature Reasoning (RFR) module to recurrently infer missing regions \cite{Li_2020_CVPR}. Although deep learning neural networks often present superior performance over traditional counterparts, most of them are trained on \emph{natural} images so that a domain transferring might be necessary before being directly applied to retinal OCT images \cite{2018Deep}. More importantly, a relatively large dataset and high computational costs are required for training and testing. 

Due to these challenges, incorporating sparse representation into medical image processing remains as a popular alternative. Fang \emph{et al.} proposed a sparsity-based tomographic method for denoising Spectral-Domain OCT volumetric data \cite{Fang:12}. Liu \emph{et al.} proposed a dictionary-based sparse representation method recently for saturation artifacts removal in OCT images \cite{2021Inpainting}. While their method works very well with saturation artifacts narrower than the patch width, it appears to be less effective when large areas of shadow are presented. 

To address these issues, we propose a multi-scale sparse representation-based shadow inpainting framework that reconstructs shadowed regions at multiple scale branches. For wide shadows, we adopt a \emph{downsampling-inpainting-upsampling} work flow to utilize the information of other scale and reduce the shadow width for inpainting retinal OCT images. Specifically, the proposed method extracts the structural information via sparse representation, and later performs fusion with the aid of a CNN-based super-resolution module. In addition, we design a regularizing module to constraint the reconstructed output while simultaneously merging information of other scales. We validate the proposed method by conducting experiments on synthetic and real-world shadows. Both qualitative and quantitative results demonstrate the superior performance of our proposed model.

\section{METHODS}

\subsection{Framework}
\begin{figure} [ht]
\begin{center}
\begin{tabular}{c}
\includegraphics[width=0.95\linewidth]{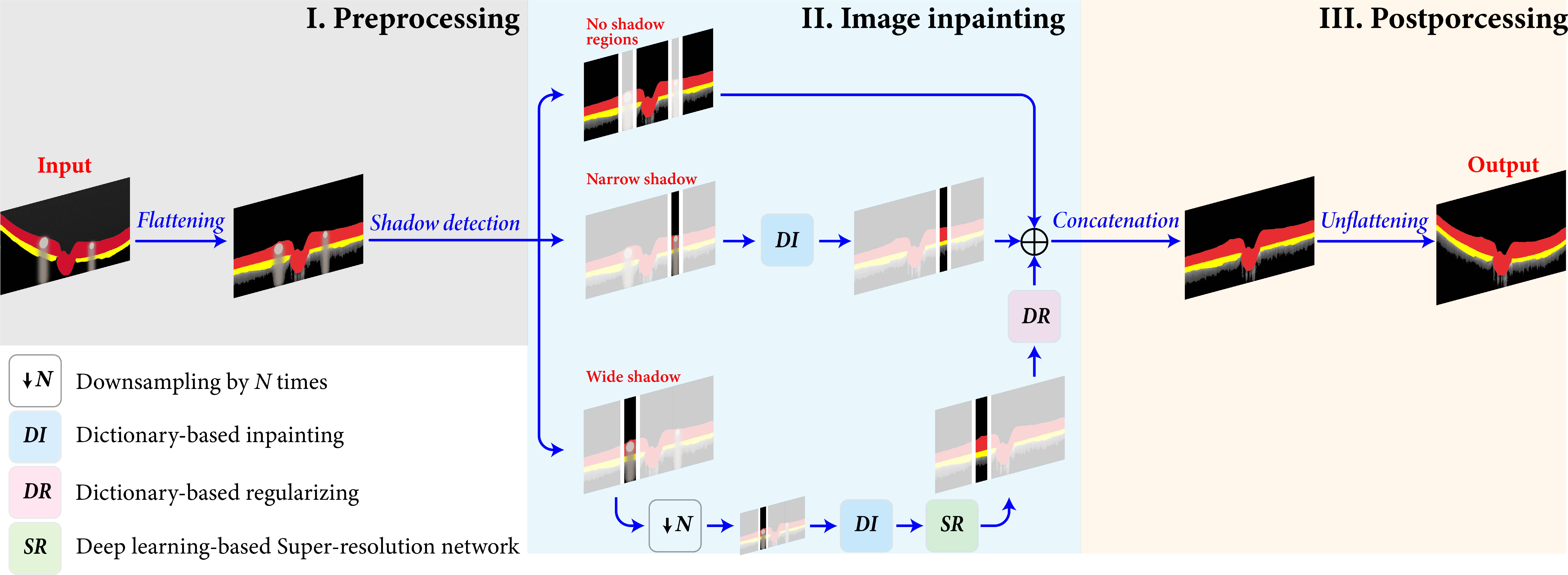}
\end{tabular}
\end{center}
\caption[framework] 
{ \label{fig:framework} 
The framework of the proposed method.}
\end{figure} 
As illustrated in Figure \ref{fig:framework}, the proposed method consists of three procedures: 1) preprocessing, 2) image inpainting, and 3) postprocessing. An input retinal OCT image that contains shadowed regions of different widths is first flattened and fed to an A-line-based shadow detection algorithm. Then, we channel shadows into three different branches based on the presence and the width of the shadows. Specifically, the flattened image is striped into the different types of strips including: (1) shadow-free regions, (2) narrowly shadowed regions and (3) widely shadowed regions. The regions with no shadow presented will be kept intact, while the narrowly shadowed regions will be processed in the dictionary-based inpainting (DI) module. 

For wide shadows, we first downsample them to decrease their widths and then send them to the DI module, where they will be inpainted based on sparse representation theory. The inpainted strips will first be upsampled back to the original resolution via an super-resolution network to restore the image quality and later be regularized by a dictionary-based regularizing (DR) module to integrate complementary feature information from a higher scales. Finally, we perform the post-processing by concatenating and unflattening the aforementioned regions to form a shadow-free output.

\subsection{Preprocessing}

The preprocessing process includes image flattening and shadow detection. We identify the Bruch’s membrane (BM), which is assumed to be the brightest layer within all retinal tissue, by picking up the pixels with the highest intensity of each A-line. Then, the BM’s location is fitted using a robust version of the locally estimated scatterplot smoothing (LOESS \cite{LOESS}) method. Considering that the pixel intensity could be perturbed by strong scatterers, we further detect the outliers in the fitting as good indicators of the presence of vessels and shadows. Besides, by moving the BM to a fixed depth according to the fitted curve with outliers removed, the A-lines are realigned to get the flattened image. Finally, a dilation operation is performed along with a connectivity test to group neighboring shadowed A-lines together and create a binary shadow mask where shadowed pixels are labeled as “0” and shadow-free pixels as “1”. To ensure that the shadows do not coincide with the image boundary, the detected edges of shadows in the binary mask are broadened by a margin of 2 pixels.

\subsection{Image Inpainting}
The image inpainting process is divided into different processing branches: no shadow  regions, narrowly shadowed regions and widely shadowed regions. The threshold used to separate shadows is set as the patch size used in the dictionary learning stage. For shadow-free regions, the image patch will be kept intact.
\subsubsection{Narrowly shadowed regions}
The narrowly shadowed regions in a flattened image are cropped based on the shadow mask and then fed to a DI module, where it is divided into overlapping patches for inpainting. These patches have a size of $a\times b$, and are vectorized as $\boldsymbol{y} \in \rm{\mathbf{R}}^{ab \times 1}$. For training the overcomplete dictionary $\mathbf{D} \in \rm{\mathbf{R}}^{ab \times M}$ ($ab < M $, where $M$ represents the number of dictionary atoms) used in the DI module, we select 33 high quality retinal OCT images $\mathbf{Y}$ without shadows from the dataset \cite{Li:21}, in which black background patches are excluded. Based on the sparsity of images, we train the dictionary as follows,
\begin{equation}
\mathop{\min}\limits_{\mathrm{D},\alpha} \| \boldsymbol{y}_t-\mathbf{D}\boldsymbol{\alpha}\|_2 \quad\mathrm{s.t.}\quad \Vert \boldsymbol{\alpha} \Vert_0 \leq L
\label{Equ: 1}
\end{equation}
where $\boldsymbol{\alpha} \in \rm{\mathbf{R}}^{M \times 1}$ is the sparse coefficient, $L$ is the sparsity level and $\boldsymbol{y_t}$ represents the shadow-free patch vector from the training dataset. It should be noted that the learned dictionary $\mathbf{D}$ is later used in DI model and DR model. Considering a case where we have shadowed pixels in a patch, we first remove the shadowed pixels from vector $\boldsymbol{y}$ and rows with the same index from the dictionary $\mathbf{D}$ to get $\boldsymbol{y}^{\prime}$ and $\mathbf{D}^{\prime}$. We assume that the remaining pixels are reliable. Sparse coefficient \cite{2016A} $\boldsymbol{\alpha}$ is calculated by: 
\begin{equation}
\boldsymbol{\hat{\alpha}} = \mathop{\arg\min}\limits_{\alpha} \| \boldsymbol{y}^{\prime}-\mathbf{D}^{\prime}\boldsymbol{\alpha}\|_2 \quad\mathrm{s.t.}\quad \Vert \boldsymbol{\alpha} \Vert_0 \leq L
\label{Equ: 2}
\end{equation}
As an inpainting process, the inpainted patch $\boldsymbol{\hat{y}}$ for testing patch $\boldsymbol{y}$ is calculated by:
\begin{equation}
    \boldsymbol{\hat{y}}=\mathbf{D}\boldsymbol{\hat{\alpha}}
\end{equation}
Finally, we aggregate all the inpainted patches to form the inpainted result.

\subsubsection{Widely shadowed regions}
The processing of widely shadowed regions follows a similar procedure except that the flattened image is first downsampled and the inpainted output is later upsampled. For the dictionary $\mathbf{D}_d$ used in the wide shadow inpainting, the training dataset is similar except that the OCT images are downsampled first before dividing into overlapping image patches. $\downarrow_N$ is the downsampling operation with a factor of $N$. Specifically, the downsampling factor $N$ is chosen depending on the statistics of the shadow width to ensure that the reduced shadow width is within the allowable range. 
\begin{equation}
    \mathbf{Y}_{d}=(\mathbf{Y})\downarrow_N
\end{equation}
For wide shadow branch, the input flattened image is first downsampled by $N$ times (\emph{e.g.} $N$ = 4 in this study) to reduce the width of wide shadow to below the threshold and simultaneously utilize the information of this scale. The downsampled image is then fed into the DI module and inpainted,
\begin{equation}
\boldsymbol{\hat{\alpha}}_d = \mathop{\arg\min}\limits_{\alpha_d} \| \boldsymbol{y}^{\prime}_d-\mathbf{D^{\prime}}_d\boldsymbol{\alpha}_d\|_2 \quad\mathrm{s.t.}\quad \Vert \boldsymbol{\alpha}_d \Vert_0 \leq L
\label{Equ: 3}
\end{equation}
The inpainted results are later upsampled by the Enhanced Deep Residual Network (EDSR)\cite{2017Enhanced} $\mathcal{U}$ afterwards to restore the high resolution information. 
\begin{equation}
    \boldsymbol{\hat{y}}^\prime=\mathcal{U}\left(\boldsymbol{\hat{y}}_d\right)=\mathcal{U}\left(\mathbf{D}_d\boldsymbol{\hat{\alpha}}_d\right)
\end{equation}
Furthermore, to constraint the generated inpainted results, a sparse representation-based regularizing module is designed to refine the contents after feature aggregation. The regularization is conducted by Equation \ref{Equ: 4} for all patches having inpainted pixels.
\begin{equation}
\boldsymbol{\hat{y}}^{\prime\prime}=\mathbf{D}\boldsymbol{\hat{\alpha}}_c
\label{Equ: 4}
\end{equation}
where $\boldsymbol{\hat{{\alpha}_c}}$ is calculated by:
\begin{equation}
\boldsymbol{\hat{\alpha}}_c =\mathop{\arg\min}\limits_{\alpha_c} \| \boldsymbol{\hat{y}}^\prime-\mathbf{D}\boldsymbol{\alpha}_c\|_2 \quad\mathrm{s.t.}\quad \Vert \boldsymbol{\alpha}_c \Vert_0 \leq L
\label{Equ: 5}
\end{equation}

The final results are concatenated by the outputs of three aforementioned branches. After unflattening, the shadows in the output image are replaced by inpainted results and the other reliable A-lines remain unchanged.

\subsection{Experimental Settings}
We evaluate our proposed method on retinal OCT images with both synthetic and real-world shadows. For synthetic shadows, we randomly generate 468 shadows of different widths ranging from 7 to 24 pixels in 9 OCT B-scans. For real-world shadows, 5 additional OCT B-scans which are not included in the training are tested. 
\subsubsection{Dictionary learning}
For dictionary learning, two dictionaries each containing 128 atoms are trained with the same parameters for different scales based on Equation (\ref{Equ: 1}) with $L=2$. The training dataset consists of 33 OCT B-scans selected from the dataset\cite{Li:21}, which are collected at the Ophthalmology Department of Shanghai General Hospital with a size of 1024$\times$992 pixels, corresponding to a field of view of 20.48 mm$\times$7.94 mm. For the dictionary of the other scale, images are downsampled by a factor of 4 correspondingly. After cropping, 10,000 overlapping patches are extracted per B-scan with a size of 8$\times$8 pixels, which makes the total 330,000 patches.
\subsubsection{Deep learning-based super-resolution network}
To train the deep learning-based super-resolution network, 800 OCT B-scans from OCTA-500 dataset \cite{2020IPN} are randomly picked for pretraining the EDSR, centered on the macula or the optic nerve. We then finetune the pretrained model by using the collected dataset\cite{Li:21} with 544 B-scans for training and 181 B-scans for validation. It should be noted that the training dataset for super-resolution network and DI module are mutually exclusive. Each B-scan is cropped into overlapping patches with a size of 48$\times$48 pixels.
\subsubsection{Deep learning-based inpainting network}
To validate the proposed method, we compared our model with existing traditional methods together with state-of-the-art deep learning-based inpainting methods. Specifically, SI, TV\cite{ipol2012}, the CNN-based network RFR\cite{Li_2020_CVPR} and RN\cite{yu2020region} are chosen. RFR and RN are pretrained in \emph{natural} image datasets. We use their original implementation and the models trained on Paris StreetView \cite{10.1145/2185520.2185597} training set provided by the authors. We then finetuned the pretrained model on OCT images dataset which contains 180,800 randomly selected images from OCTA-500 and 11,208 images from our training dataset. Specifically, the training images include missing regions, which are generated by applying synthetic binary shadow masks with a normally distributed mask width. The images and the masks are then cropped into patches with a size of 256$\times$256 to conform the input format of the networks with no further resizing.
\subsubsection{Segmentation network}
To evaluate the inpainting performance of different methods on machine analysis, we fed the inpainted image to a widely used segmentation network UNet \cite{10.1007/978-3-319-24574-4_28}. The Dice score of the segmented results are calculated to analyze the performance of different inpainting methods.

\section{RESULTS}
\begin{figure} [ht]
\begin{center}
\includegraphics[height=7.4cm]{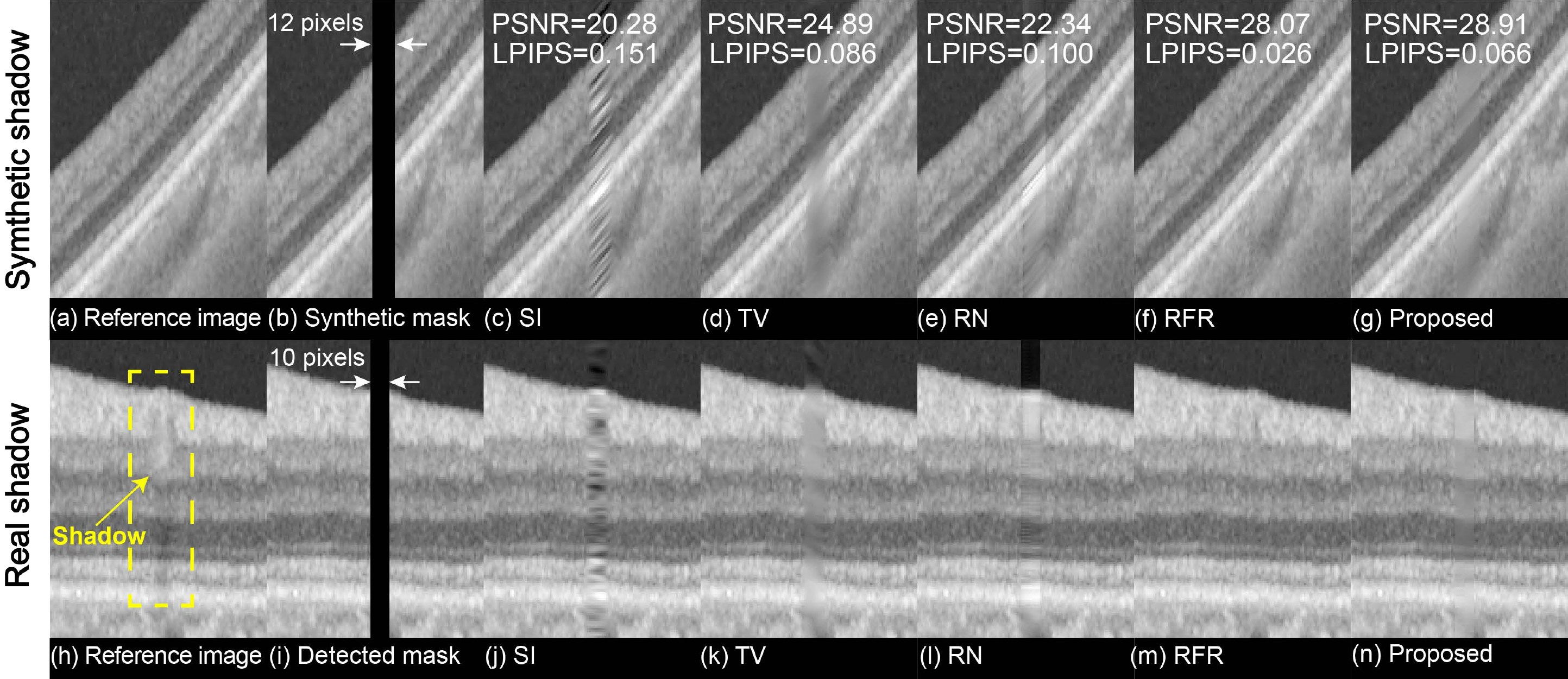}
\end{center}
\caption{Visual comparison between different methods for synthetic shadow and real shadow.}
\label{fig:result1} 
\end{figure}
A qualitative comparison between the results obtained by different techniques for synthetic and real-word shadows is shown in Figure \ref{fig:result1}. The black vertical bars shown in Figure \ref{fig:result1}(b) and (i) are the synthetic and the detected shadow mask with a shadow width of 12 pixels and 10 pixels, respectively. In Figure \ref{fig:result1}(c) and (j), the shadows inpainted by SI show severe discontinuities within each layer. On the other hand, TV generates a roughly smooth structure. However, the results still contain blurs and over-smoothing textures inside the shadow region as shown in Figure \ref{fig:result1}(d) and (k). For CNN-based RN network, the resultant patches appear more realistic and less blurry as given in Figure \ref{fig:result1}(e) and (l). However, it generates some unexpected contrast artifacts which lead to a very low PSNR and SSIM. Another deep learning-based network RFR generates the most visually pleasing results as presented in Figure \ref{fig:result1}(f) and (m) while the inpainted images manifest the lowest LPIPS as well. Nonetheless, it possesses a slightly lower PSNR than our proposed method shown in Figure \ref{fig:result1}(g) and (n). Nonetheless, the visual performance of our proposed method preserves layer continuity but is not as good as RFR. Nonetheless, the proposed method has the highest PSNR and SSIM, but a higher LPIPS than RFR.

To quantitatively evaluate the inpainting quality, the mean and standard deviation of PSNR, LPIPS and SSIM are calculated for different techniques on the synthetic testing set. Table \ref{tab: table} shows the results with best performance value written in bold. It can be observed that our model outperforms other methods in both PSNR and SSIM. The efficacy of the proposed method could partially attribute to the sparsity-promoting loss function in DI module and the proposed multi-scale scheme. As to the LPIPS metric, the proposed method underperforms the RFR network, which is largely consistent with the visual evaluation mentioned above.
\begin{table}[ht]
\caption{Objective quantitative metrics and running time compared with other methods on synthetic shadows.} 
\label{tab: table}
\begin{center}       
\begin{tabular}{|l|c|c|c|c|c|}
\hline
\rule[-1ex]{0pt}{3.5ex}  Metric & SI & TV & RN & RFR & Proposed  \\
\hline
\rule[-1ex]{0pt}{3.5ex}  PSNR (dB)$\uparrow$ & 21.12$\pm 2.38 $ & 24.04$\pm 4.13 $ & 17.91$\pm 2.09 $ & 29.99$\pm 0.97 $ & \textbf{31.21$\boldsymbol{\pm 1.08}$}   \\
\hline
\rule[-1ex]{0pt}{3.5ex}  SSIM$\uparrow$ & 0.30$\pm 0.11 $ & 0.59$\pm 0.08 $ & 0.47$\pm 0.08 $  & 0.69$\pm 0.07 $ & \textbf{0.79$ \boldsymbol{\pm 0.06} $}  \\
\hline
\rule[-1ex]{0pt}{3.5ex}  LPIPS$\downarrow$ & 0.306$\pm 0.034 $ & 0.245$\pm 0.047 $ & 0.249$\pm 0.026$ & \textbf{0.064$\boldsymbol{\pm 0.018} $} & 0.170$\pm 0.063 $  \\
\hline
\rule[-1ex]{0pt}{3.5ex} Time(Train$+$Test) (s) & 0$+$139 & 0$+$6,226 & 71,200$+$8,236 & 128,400$+$977 & 4,206$+$1,203\\  
\hline
\end{tabular}
\end{center}
\end{table}

\begin{figure}[ht]
  \centering
  \includegraphics[height = 4.4cm]{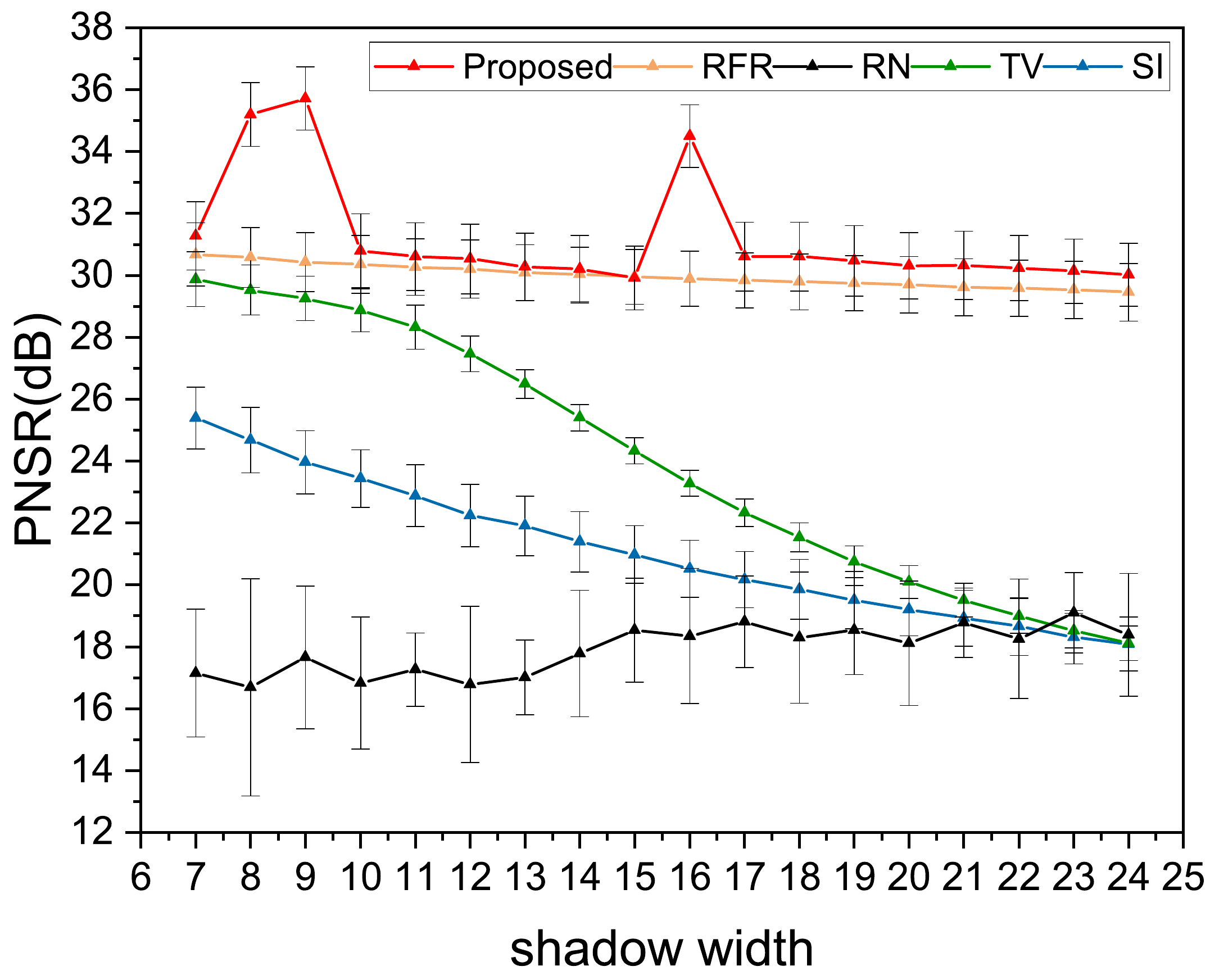}
  \includegraphics[height = 4.4cm]{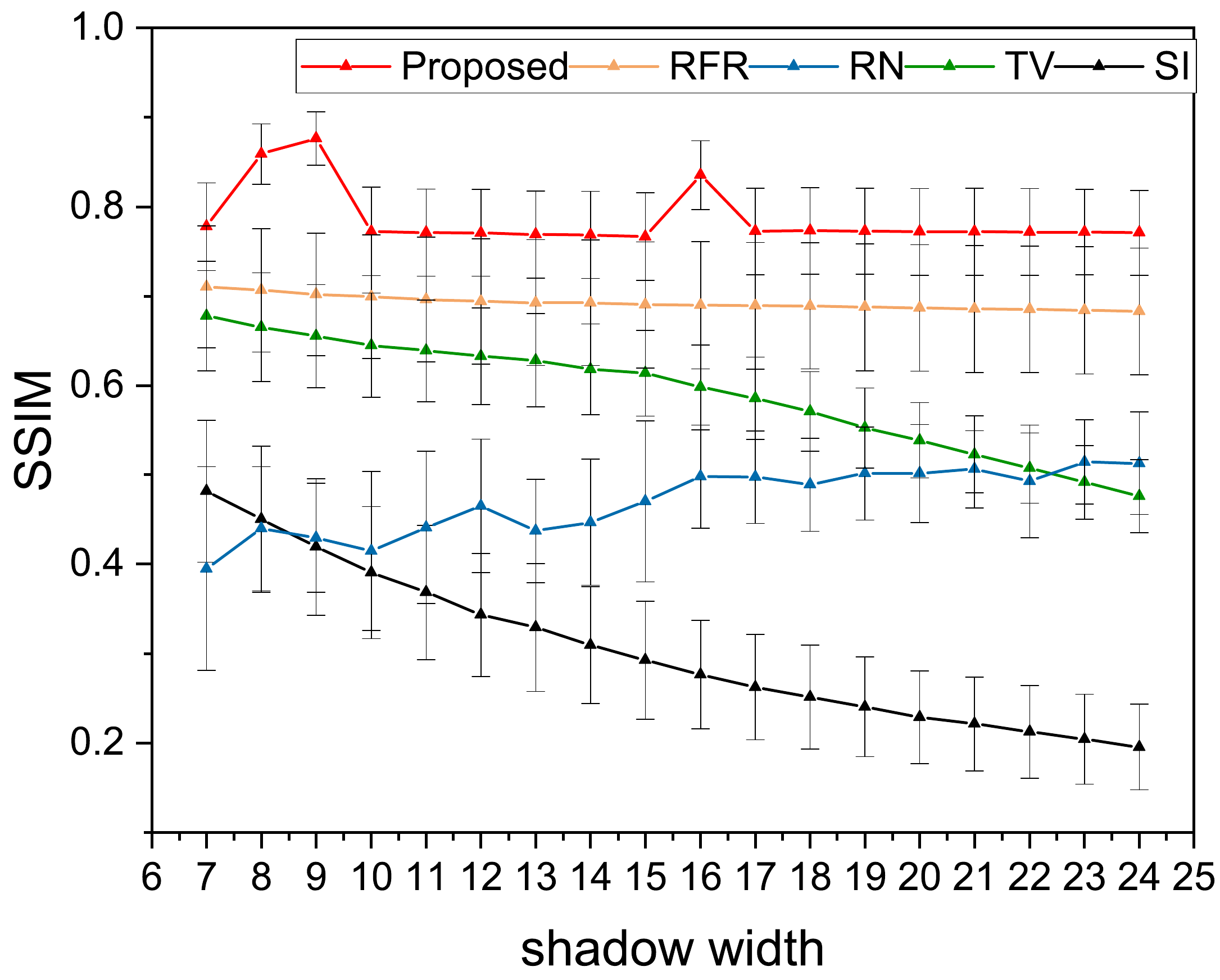}
  \includegraphics[height = 4.4cm]{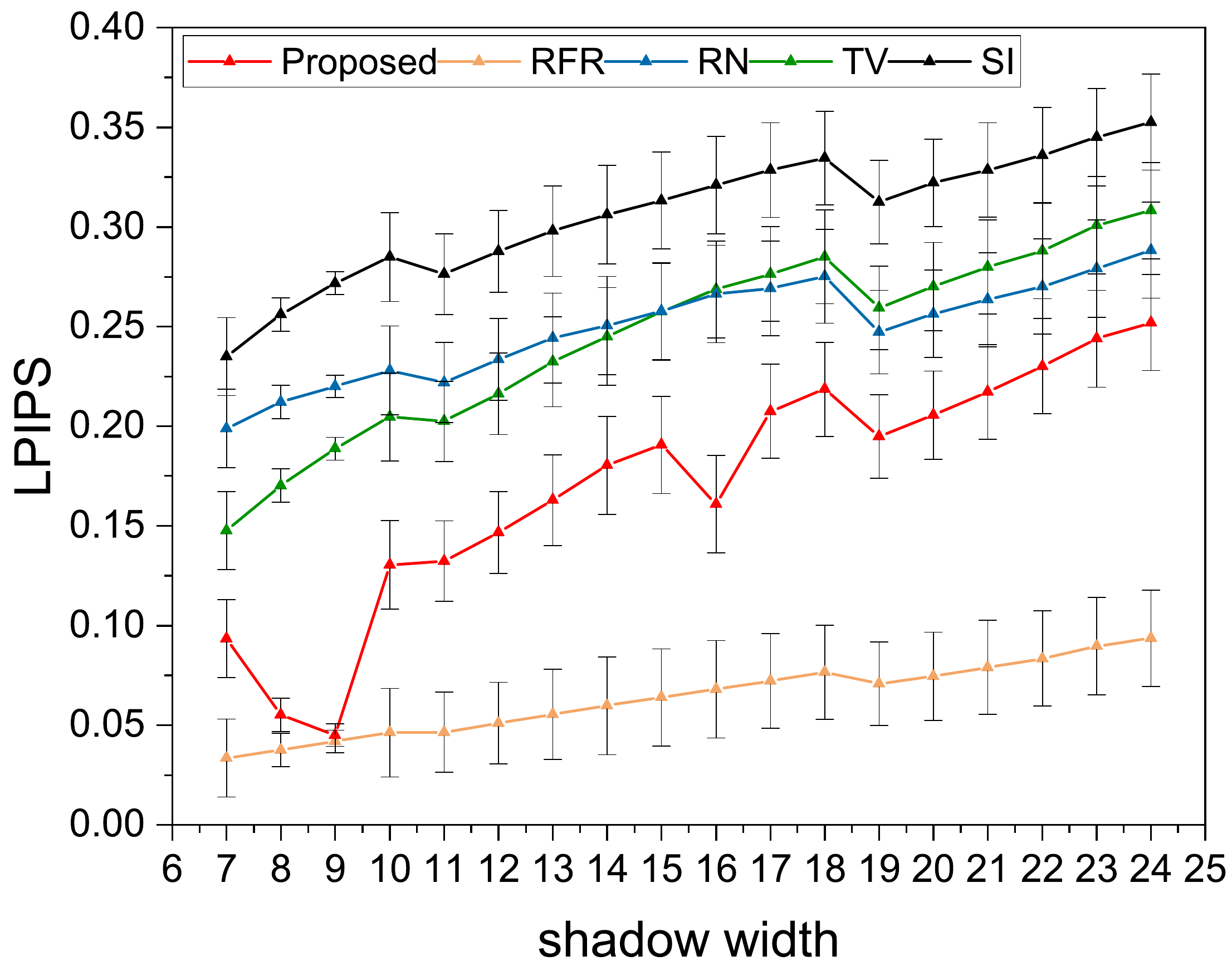}
  \caption{PSNR, SSIM and LPIPS curve of different methods over shadow width.}
\label{fig: results}
\end{figure}

To further evaluate the capability of the proposed technique on handling wide shadows, we repeat the inpainting experiment on various shadow widths (ranging from 7 pixels to 24 pixels) by using different techniques and calculating the corresponding PSNR, LPIPS, and SSIM. The results are plotted in Figure \ref{fig: results}. One of the major advantages of the proposed method is that the inpainting performance possesses a relatively lower dependency on the shadow width. As shown in Figure \ref{fig: results}, the proposed method and RFR remains almost constant for PSNR and SSIM, while the performance of other traditional methods drops dramatically as the shadow width increases (compared with the patch size). We suggest it is because the proposed method utilizes a multi-scale framework to handle the wide shadows. Instead of being directly inpainted, the wide shadows are first narrowed by downsampling to ensure a satisfactory inpainting performance at a reduced resolution. The following procedures including a deep learning-based upsampling and a dictionary-based regularization minimize the impairment brought by the downsampling to the resolution. Since the patch size and threshold are set to be 8 pixels, there are performance improved when the width of shadow reaches 8 pixels and 16 pixels. Furthermore, the LPIPS for RFR is the best while our method has the similar trend with traditional methods.

\begin{figure}[ht]
\centering
\includegraphics[height = 8.5cm]{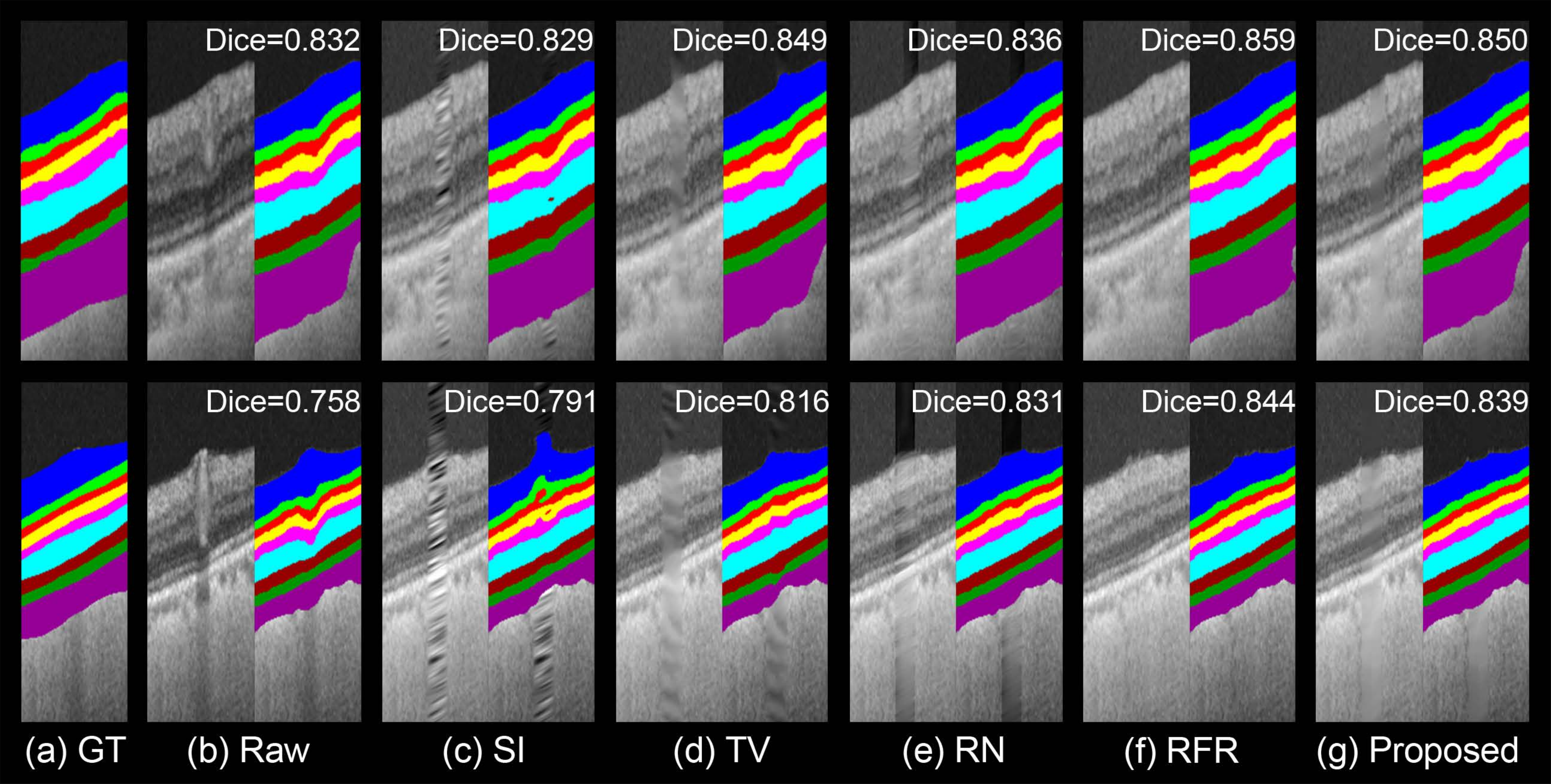}
\caption{Retinal OCT image segmentation results before and after inpainting.}
\label{fig: Segmentation}
\end{figure}

We further compare the effectiveness of shadow inpainting for machine diagnosis. Specifically, raw OCT images containing real shadows and the corresponding inpainted results by different methods are sent into a retinal OCT image segmentation network UNet for retinal layers segmentation. The results are provided in Figure \ref{fig: Segmentation} with nine layers annotated by different colors. The Dice score between ground truth and prediction is computed for the quantitative evaluation of segmentation performance. Due to the presence of the large blood vessels, the segmentation results obtained from raw images suffer from obvious distortion as shown in Figure \ref{fig: Segmentation}. From Figure \ref{fig: Segmentation} (c), the poor inpainted results of SI causes segmentation results to create scattered labels and layer discontinuities emerge. Different from SI, TV and RN decrease the scattered labels but still exists bending of retinal layers influenced by less satisfactory inpainted results. As a comparison, the proposed method achieves a better Dice score and less visual error except for the choroid layer. Finally, it is worth noting that RFR achieves the best performance and has the highest Dice score, which is consistent with the above analysis.

To sum up, our proposed multi-scale framework can achieve expected inpainting results and has slightly higher PSNR and SSIM metrics, while the deep learning method RFR achieves the best visual performance and LPIPS score.

\section{DISCUSSION}
\subsection{Time efficiency}
The last row of Table \ref{tab: table} lists the training and testing time of different methods. Traditional methods like TV need to iteratively propagate information from the neighboring regions to the missing area, which leads to a relatively longer testing time compared to that of SI. As for deep learning-based methods, they spend most of the time on training the network parameters. In contrast, the training time for the proposed method is much smaller than deep neural network, since the dictionary learning procedure does not include the computational-heavy convolution operations in the data-driven deep learning network. Besides, it should be noted that the downsample and upsample operation add additional time to the test procedure.
\subsection{Performance of deep learning network}
In this paper, we select two deep learning-based networks for comparison: RN and RFR. RN adopts an encoder-decoder framework for solving the mean and variance shift problem and improving the training. RFR recurrently infers the missing boundaries using the convolutional feature maps. They are both widely used \emph{natural} image inpainting networks. When deep learning networks work properly, they present a better performance than most traditional methods and sparse representation-based methods. However, there is a chance that some deep learning networks might overfit the data and breakdown during the testing.

For RN network, the inpainting results for OCT images are not good as we expected. RN is designed and pretrained on \emph{natural} images, which have the different characteristics of \emph{medical} images. It might be necessary to adjust the network architecture or use a transferring network in future work. 

For RFR network, the results has the most realistic visual appearance and the lowest LPIPS value, which indicates its powerful performance. RFR can capture more high-level structure features and shows promising results for images with distinct patterns. Though the RFR has a smaller PSNR, we conjecture that it captures the noise feature of coherent noise which has a Poisson distribution. Thus, in the inpainting process, RFR may add some unexpected noise into the inpainted region while our method manifests an over smoothing effect. 

\section{CONCLUSIONS}
In conclusion, we propose a novel multi-scale sparse representation-based shadow inpainting framework for retinal OCT images. A \emph{downsampling-inpainting-upsampling} work flow is employed to handle wide shadows cast by large vessels. Moreover, a CNN-based super-resolution network is implemented for upsampling and simultaneously enhancing spatial resolution, which minimizes the penalties imposed by the downsampling procedure. Experimental results obtained by our methods outperform other existing methods on both the synthetic and real-world shadow inpainting tasks. In the future, this proposed framework will be extended to other medical image inpainting or reconstruction processing tasks by exploring structure characteristics in medical images through sparse representation.

\acknowledgments 
 
The authors would like to thank Zhenxing Dong for his contribution to this paper. This work is supported in part by National Natural Science Foundation of China (61905141, 62035016) and Shanghai Sailing Program (19YF1439700). 


\bibliographystyle{spiebib} 

\end{document}